\documentclass[a4paper,fleqn]{cas-sc}

\usepackage[authoryear]{natbib}
\usepackage{amsmath}
\usepackage[normalem]{ulem}
\usepackage{graphicx}
\usepackage{float}
\usepackage{microtype}
\usepackage{hyperref}
\usepackage{subcaption}
\hypersetup{
    colorlinks=false,
    citecolor=blue,
    linkcolor=blue,
    urlcolor=blue
}

\def\tsc#1{\csdef{#1}{\textsc{\lowercase{#1}}\xspace}}
\tsc{WGM}
\tsc{QE}
\tsc{EP}
\tsc{PMS}
\tsc{BEC}
\tsc{DE}

\begin{document}
\let\WriteBookmarks\relax
\def\floatpagepagefraction{1}
\def\textpagefraction{.001}
\shorttitle{Alzheimer's Disease Continuum}
\shortauthors{Y. Zhang et~al.}

\title [mode = title]{Longitudinal Bayesian Learning of Continuous Disease Position across the Alzheimer's Disease Continuum}

\author[1]{Yingying Zhang}[bioid=1, orcid=0009-0009-7899-5902]
\ead{yingying.zhang01@utrgv.edu}
\author[2]{Kun Zhao}[bioid=2]
\author[3]{Guodong Liu}[bioid=3]
\author[4]{Qi Huang}[bioid=4]
\author[1]{Pengfei Gu}[bioid=5]
\author[1]{Dongchul Kim}[bioid=6]
\author[1]{Erik Enriquez}[bioid=7]
\author[5]{Alex D. Leow}[bioid=8]
\author[6]{Paul M. Thompson}[bioid=9]
\author[7]{Heng Huang}[bioid=10]
\author[8]{Hongchang Gao}[bioid=11]
\author[2]{Liang Zhan}[bioid=12]
\author[1]{Haoteng Tang}[bioid=13, orcid=0000-0003-0323-1755]
\cormark[1]
\ead{tanghaoteng@gmail.com}
\author[]{for the Alzheimer's Disease Neuroimaging Initiative (ADNI) Project}

\credit{Conceptualization of this study, Methodology, Software}

\affiliation[1]{organization={Department of Computer Science, University of Texas Rio Grande Valley},
                addressline={1201 W. University Dr.}, 
                city={Edinburg},
                postcode={78539}, 
                state={TX},
                country={USA}}
\affiliation[2]{organization={Department of Electrical \& Computer Engineering, University of Pittsburgh},
                addressline={3700 O'Hara Street}, 
                city={Pittsburgh},
                postcode={15261}, 
                state={PA},
                country={USA}}
\affiliation[3]{organization={Eli and Lilly Company},
                addressline={Lilly Corporate Center}, 
                city={Indianapolis},
                postcode={46285}, 
                state={IN},
                country={USA}}
\affiliation[4]{organization={Department of Radiology, Washington University in St. Louis},
                addressline={4525 Scott Ave}, 
                city={St. Louis},
                postcode={63110}, 
                state={MO},
                country={USA}}
\affiliation[5]{organization={Department of Psychiatry, University of Illinois Chicago},
                addressline={1601 W. Taylor St.}, 
                city={Chicago},
                postcode={60612}, 
                state={IL},
                country={USA}}
\affiliation[6]{organization={Imaging Genetics Center, University of Southern California},
                addressline={4676 Admiralty Way}, 
                city={Marina Del Rey},
                postcode={90292}, 
                state={CA},
                country={USA}}
\affiliation[7]{organization={Department of Computer Science, University of Maryland},
                addressline={8125 Paint Branch Dr.}, 
                city={College Park},
                postcode={20742}, 
                state={MD},
                country={USA}}
\affiliation[8]{organization={Department of Computer and Information Sciences, Temple University},
                addressline={1925 N. 12th Street}, 
                city={Philadelphia},
                postcode={19122}, 
                state={PA},
                country={USA}}

\cortext[cor1]{Corresponding author}

\begin{abstract}
Alzheimer's disease (AD) progresses as a continuous biological process, whereas most existing neuroimaging-based artificial intelligence methods remain limited to discrete diagnosis or clinical score prediction from cross-sectional imaging. In this work, we propose Disease Continuum Positioning (DCP), a longitudinal Bayesian Learning framework that continuously estimates disease severity from longitudinal diffusion tensor imaging (DTI). Specifically, DCP models disease severity as a low-dimensional probabilistic latent variable by jointly integrating longitudinal observations with weak clinical supervision, from which the proposed Disease Continuum Score (DCS) is derived to quantify an individual's position along the Alzheimer's disease continuum together with its associated uncertainty. Extensive experiments on the Alzheimer's Disease Neuroimaging Initiative (ADNI) cohort demonstrate that DCP consistently outperforms representative disease progression methods. More importantly, comprehensive validation analyses show that DCS accurately characterizes disease severity, exhibits strong clinical relevance, preserves longitudinal disease evolution, and predicts future disease conversion. These results suggest that DCS provides a quantitative imaging-derived representation for continuous assessment of Alzheimer's disease progression beyond conventional diagnostic labels and clinical scores.
\end{abstract}

\begin{graphicalabstract}
\centering
\includegraphics[width=\textwidth]{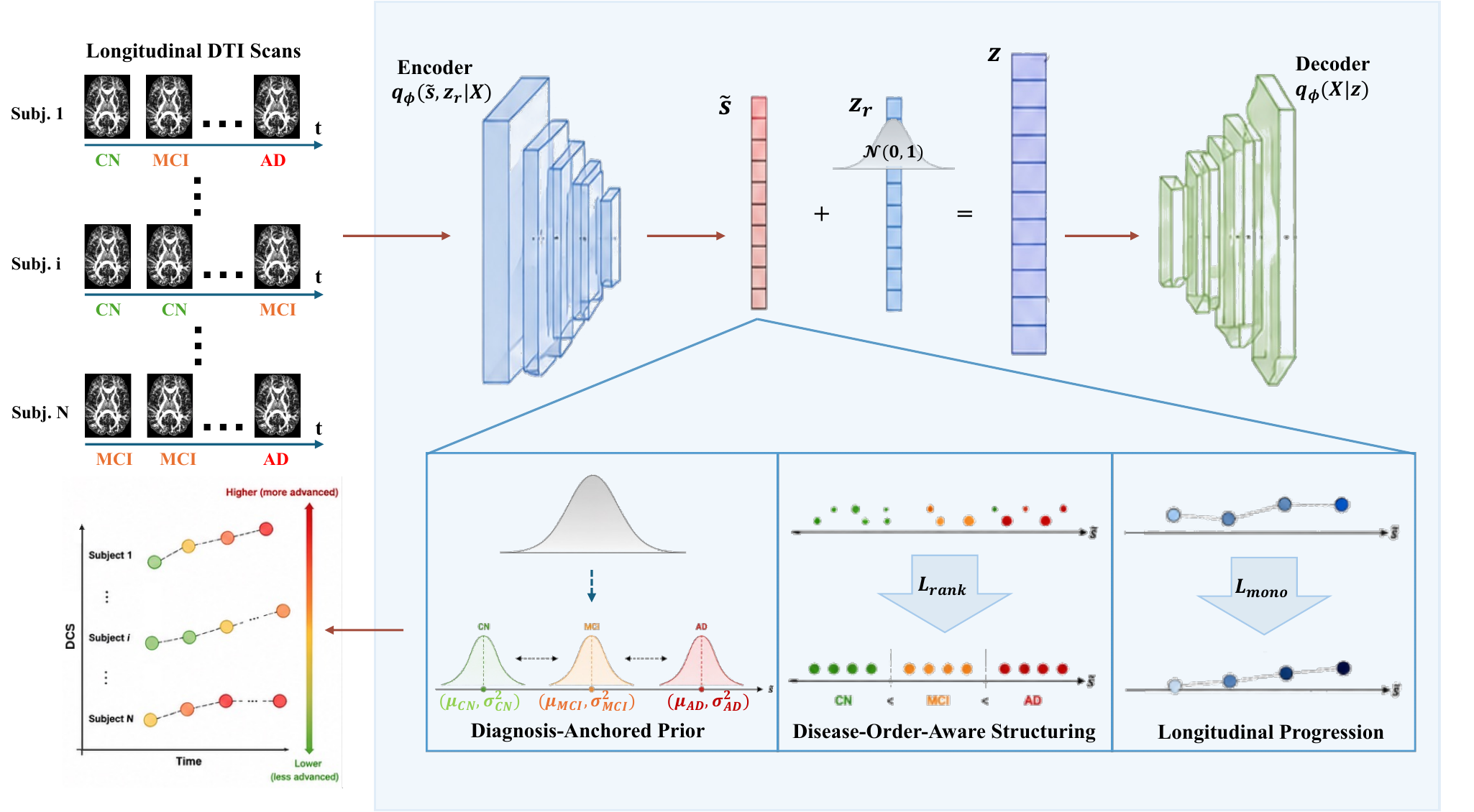}
\end{graphicalabstract}

\begin{highlights}
\item We propose Disease Continuum Positioning (DCP), a new framework that generates an uncertainty-aware Disease Continuum Score (DCS) for continuous patient positioning along the Alzheimer’s disease continuum.
\item DCP explicitly learns disease position beyond discrete diagnostic stages, enabling fine-grained characterization of individual disease severity.
\item DCS captures clinically meaningful disease variation and provides prospective information on future CN-to-MCI and MCI-to-AD conversion.
\end{highlights}


\maketitle

\section{Introduction}
Alzheimer's disease (AD) is one of the leading causes of dementia worldwide, imposing substantial medical, social, and economic burdens on patients and healthcare systems \citep{chen2024global,livingston2020dementia}.
As the global population ages, the prevalence of AD continues to increase, emphasizing the importance of accurately characterizing disease progression for early diagnosis, prognosis, and disease management.
Diffusion tensor imaging (DTI) is a non-invasive magnetic resonance imaging (MRI) technique that quantifies the microstructural integrity of white matter by measuring water diffusion within brain tissues. 
Compared with conventional structural MRI, DTI is more sensitive to subtle white matter abnormalities that may occur before significant brain atrophy becomes apparent \citep{le2003looking,o2011introduction}. 
Previous studies have shown that diffusion-derived measures, including fractional anisotropy (FA), mean diffusivity (MD), radial diffusivity (RD), and axial diffusivity (AD), are closely associated with AD progression and cognitive decline \citep{amlien2014diffusion,chen2023abnormal}. 
These advantages make DTI a promising imaging modality for characterizing the continuous progression of Alzheimer's disease.

Recent advances in deep learning have substantially advanced the analysis of neuroimaging data for AD. 
Most existing studies formulate disease assessment as either a classification task \citep{zhan2014multiple,zhan2015boosting,jiang2024anatomy,marwa2023mri,cai2022graph,wen2020convolutional,kunanbayev2024training,tang2023signed,qiu2022multimodal,alp2024joint,zhao2025r-genima}, aiming to assign subjects to discrete clinical stages (e.g., cognitively normal or CN, mild cognitive impairment or MCI, and AD); or a regression task \citep{tabarestani2020distributed,huang2025model,bass2022icam,tang2022hierarchical,tang2022hierarchicalfrontiers,tang2022contrastive}, predicting cognitive or clinical assessments such as MMSE or ADAS-Cog from neuroimaging data. 
These paradigms have achieved remarkable success in automated diagnosis and disease assessment.
Nevertheless, both formulations provide only partial descriptions of the underlying disease process. 
Brain dementia progresses as a continuous biological process, with pathological changes accumulating gradually over many years before crossing diagnostic boundaries \citep{thompson2007tracking,jack2013tracking,jagust2018imaging,jack2018nia}.
Consequently, patients assigned to the same clinical stage may exhibit substantially different degrees of neurodegeneration and future progression risk, whereas patients from adjacent stages may share highly similar pathological characteristics. 
Likewise, although clinical assessments provide more details of disease evolvement than categorical diagnoses, they remain rating scales rather than truly continuous representations of disease progression. 
Moreover, these scores primarily reflect observable cognitive performance and are subject to measurement variability, making them imperfect surrogates of the underlying disease process.
These limitations motivate the development of continuous disease representations derived directly from neuroimaging data. 
Rather than assigning subjects to several isolated diagnostic categories, a continuous disease score provides a quantitative position along the disease continuum, enabling finer characterization of disease severity and more sensitive assessment of individualized disease progression.

Although continuous disease representations provide a finer characterization of disease severity than discrete diagnostic labels, learning such representations solely from cross-sectional neuroimaging remains fundamentally challenging. 
Cross-sectional imaging captures only a single observation of each subject and therefore lacks information about the temporal evolution of disease. 
Consequently, observed differences between subjects may arise from either distinct stages of neurodegeneration or inherent inter-subject anatomical variability, making these factors difficult to disentangle from a single imaging snapshot.
Longitudinal neuroimaging alleviates this limitation by repeatedly acquiring images from the same individual, thereby enabling direct observation of within-subject structural changes over time rather than relying solely on inter-subject variability~\citep{ouyang2022longitudinal,fischer2025longitudinal}. 
More importantly, repeated observations naturally establish temporal ordering among visits, providing explicit constraints on disease evolution that are fundamentally unavailable in cross-sectional studies~\citep{zhao2021longitudinal,ouyang2021self}.
Rather than relying exclusively on static population differences, disease representations can be constrained to evolve consistently along each individual's longitudinal trajectory. 
Such temporal constraints facilitate the learning of continuous disease representations that are both biologically meaningful and temporally consistent, thereby providing a more faithful characterization of disease progression.

Substantial efforts have been devoted to modeling AD progression, however, existing computational frameworks remain limited in their ability to learn an interpretable and continuous representation of disease severity directly from neuroimaging.
Classical disease progression models, such as the Event-Based Model (EBM) \citep{fonteijn2012event} and the SuStaIn framework \cite{young2018sustain}, estimate disease progression by inferring event sequences or ordinal disease stages from population-level biomarker evolution \citep{tandon2024ssustain,dejonge2026disease}. 
While these models effectively recover disease ordering at the population level, they inherently represent progression using discrete stages, limiting their ability to characterize fine-grained differences among individuals within the same stage.
Supervised learning methods instead learn continuous outputs by directly predicting cognitive or clinical measurements, such as MMSE or ADAS-Cog~\citep{feng2022deep,natureaging2026}. 
However, these approaches optimize prediction of clinical observations rather than the underlying disease progression itself, making the learned representation dependent on downstream clinical assessments instead of intrinsic disease severity.
More recently, longitudinal representation learning methods have exploited repeated neuroimaging through self-supervised or weakly supervised learning to learn temporally consistent latent representations~\citep{zhao2021longitudinal,ouyang2021self,ouyang2022longitudinal}. 
Although these methods successfully capture longitudinal consistency, the learned latent representations remain abstract feature embeddings without an explicitly defined disease severity coordinate, limiting their interpretability and clinical utility.
A principled framework is therefore needed to directly model disease severity as a continuous and interpretable coordinate, rather than discrete stages, clinical observations, or unconstrained latent embeddings.

Motivated by these observations, we hypothesize that longitudinal DTI captures rich structural information associated with Alzheimer's disease progression, from which a dedicated low-dimensional latent coordinate can be learned to continuously represent disease severity along the Alzheimer's disease continuum. 
Since disease severity is an inherently unobservable latent variable, representing it as a deterministic scalar inevitably ignores the intrinsic uncertainty arising from biological heterogeneity, imaging variability, and the partial observability of the underlying disease process. 
We therefore propose a Bayesian framework, termed \uline{Di}sease \uline{C}ontinuum \uline{P}ositioning (DCP), in which disease severity is modeled as a probabilistic latent variable rather than a deterministic value. 
Weak clinical supervision anchors the latent coordinate to the clinical ordering of disease severity, while longitudinal observations enforce temporal consistency across repeated visits. 
Given these complementary constraints, Bayesian learning infers the posterior distribution of the latent disease coordinate, yielding the most plausible continuous disease representation together with its associated uncertainty. 
The resulting probabilistic disease score enables individualized disease severity estimation and uncertainty quantification, providing a reliable framework for continuous characterization and longitudinal assessment of AD progression.
Our main contributions of this work are summarized as follows:

\begin{itemize}
\item We propose \textbf{Disease Continuum Positioning (DCP)}, a Bayesian longitudinal learning framework that jointly integrates longitudinal DTI with weak clinical supervision to infer the posterior distribution of disease severity, enabling probabilistic disease severity estimation together with uncertainty quantification.

\item We introduce the \uline{D}isease \uline{C}ontinuum \uline{S}core (DCS), a low-dimensional continuous disease severity representation estimated by DCP that quantitatively characterizes an individual's position along the Alzheimer's disease continuum beyond conventional discrete diagnostic labels and clinical scores.

\item Extensive experiments demonstrate that the proposed framework consistently outperforms state-of-the-art methods in learning disease severity representations. Comprehensive validation further shows that DCS accurately characterizes disease progression, preserves longitudinal consistency, exhibits strong clinical relevance, and facilitates future disease conversion prediction.

\end{itemize}

\section{Related Work}
\subsection{Modeling Alzheimer's Disease Progression}
Deep learning has become a dominant paradigm for Alzheimer's disease (AD) assessment from neuroimaging. 
Most existing studies formulate disease assessment as a supervised prediction problem by learning mappings from neuroimaging to either discrete diagnostic labels or continuous clinical measurements. 
Representative classification methods distinguish subjects across different disease stages using MRI or multimodal neuroimaging~\citep{qiu2022multimodal,alp2024joint}, whereas regression-based approaches directly predict disease-related measurements, such as cognitive scores or imaging-derived severity scores, from a single baseline scan~\citep{feng2022deep,natureaging2026}. 
Although these methods have demonstrated strong predictive performance, they are optimized to reproduce diagnostic labels or clinical assessments rather than learning the underlying disease severity itself. 
Moreover, most of these approaches rely on cross-sectional imaging and therefore cannot explicitly model the continuous and longitudinal nature of disease progression.
Another line of research focuses on disease progression modeling. 
Classical data-driven models, including the Event-Based Model (EBM)~\citep{fonteijn2012event} and the Subtype and Stage Inference (SuStaIn) framework~\citep{young2018sustain}, estimate disease evolution by recovering event sequences or disease stages from population-level biomarker trajectories. 
Subsequent studies further improved these frameworks by enhancing scalability and accommodating heterogeneous biomarker types~\citep{tandon2024ssustain,dejonge2026disease}. 
While these methods successfully characterize disease progression at the population level, they primarily rely on handcrafted imaging biomarkers and estimate discrete disease stages, limiting their ability to directly infer individualized continuous disease severity from high-dimensional longitudinal neuroimaging.
Different from the above approaches, our work directly learns an imaging-derived continuous disease severity representation from longitudinal DTI. 
By jointly exploiting longitudinal observations, weak clinical supervision, and Bayesian probabilistic modeling, the proposed framework estimates both individualized disease severity and its associated uncertainty, providing a quantitative representation of disease progression beyond discrete diagnosis and conventional clinical assessments.

\subsection{Longitudinal Representation Learning}
Longitudinal representation learning has recently emerged as an effective strategy for exploiting repeated neuroimaging observations beyond static diagnostic labels. Early studies primarily focused on learning temporally consistent latent representations from repeated imaging. Longitudinal Self-Supervised Learning (LSSL) enforces consistent latent transformations between consecutive visits of the same subject~\citep{zhao2021longitudinal}, while Longitudinal Neighborhood Embedding (LNE) further preserves temporal consistency by exploiting neighborhood relationships through dynamically constructed graphs~\citep{ouyang2021self}.
Subsequent studies extended this idea by disentangling disease progression from nuisance variations. Longitudinal self-supervision has been employed to separate disease-related progression from inter-subject anatomical variability~\citep{couronne2021longitudinal}, and weakly supervised disentanglement has been introduced to distinguish normal aging from disease progression through orthogonal latent representations~\citep{ouyang2022longitudinal}.
Despite these advances, the learned representations remain generic latent embeddings or latent directions whose semantics are implicitly determined by temporal consistency or disentanglement objectives, rather than explicitly representing disease severity. In contrast, the proposed DCP framework explicitly models disease severity as a dedicated low-dimensional probabilistic latent coordinate, from which the DCS is estimated under longitudinal observations and weak clinical supervision.

\begin{figure*}[h]
    \centering
    \includegraphics[width=\textwidth]{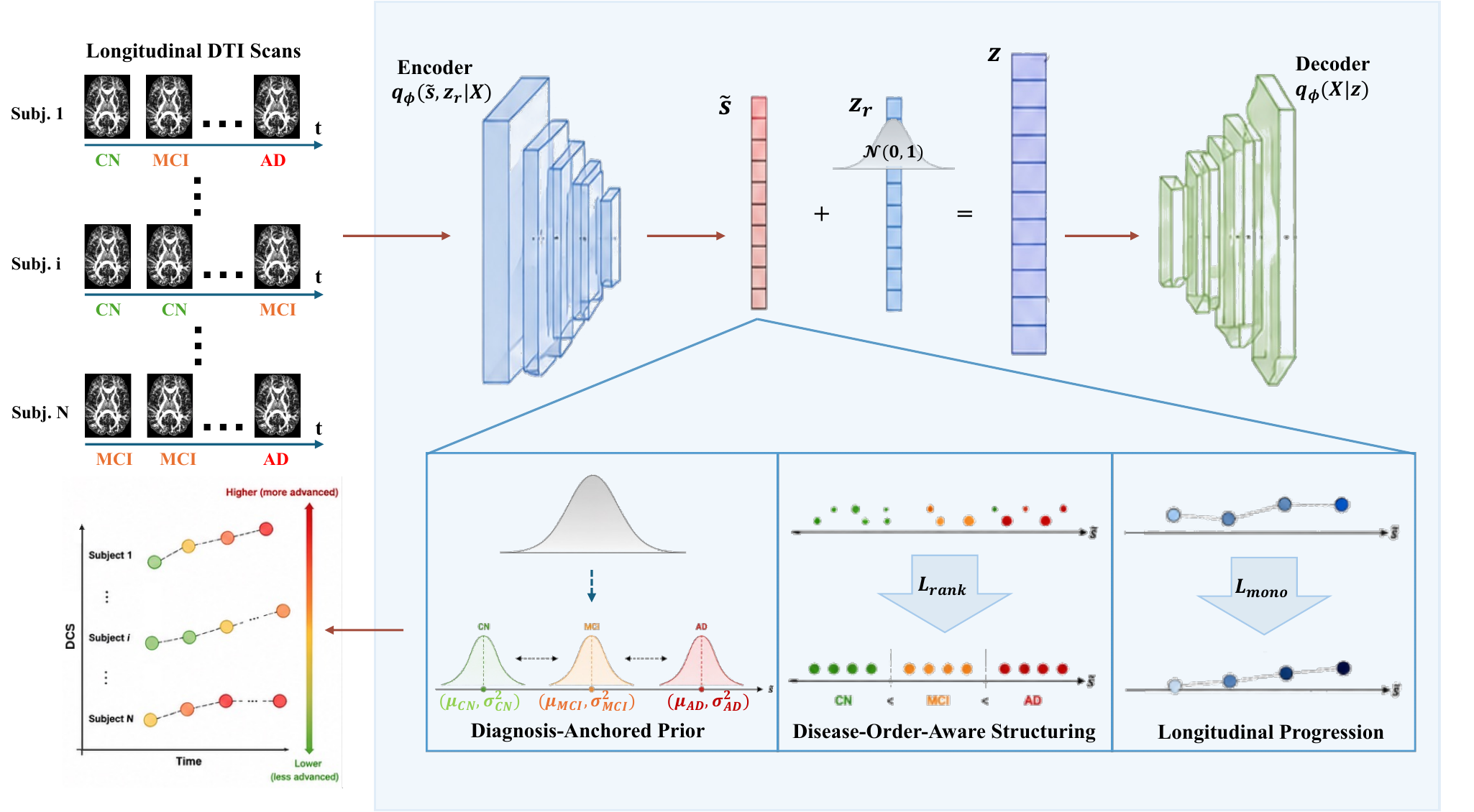}
    \caption{Overview of the proposed DCP
framework. A 3D encoder factors each longitudinal DTI scan into a
one-dimensional disease-order-aware latent $\tilde{s}$ and a residual $z_r$, from
which a decoder reconstructs the input; a diagnosis-anchored prior, an ordinal
ranking loss $\mathcal{L}_{rank}$, and a soft monotonicity loss
$\mathcal{L}_{mono}$ jointly position $\tilde{s}$ on a continuous,
diagnosis-ordered severity axis. The reported DCS, $s=\mathrm{Softplus}(\tilde{s})$, is uncertainty-aware, with higher values
indicating more advanced disease.}
    \label{fig:framework}
\end{figure*}

\section{Methodology}
\subsection{Problem Formulation}
Let \(\mathcal{D}=\{(X_i^t,y_i^t)\}\) denote a longitudinal neuroimaging dataset, where \(X_i^t\in\mathbb{R}^{C\times D\times H\times W}\) represents the neuroimaging scan acquired from subject \(i\) at time \(t\), and \(y_i^t\in\{1,2,3\}\) denotes the corresponding clinical diagnostic stage, ordered as cognitively normal (CN), mild cognitive impairment (MCI), and Alzheimer's disease (AD), respectively. Here, \(C\) denotes the number of imaging channels, and \(D\), \(H\), and \(W\) denote the spatial dimensions.
Our objective is to estimate the Disease Continuum Score (DCS), denoted by the continuous latent variable \(s_i^t\in\mathbb{R}^{+}\), which quantitatively characterizes the subject's imaging-derived disease severity along the Alzheimer's disease continuum.
The diagnostic labels provide only coarse ordinal information about disease stages and therefore cannot determine individualized disease severity within each clinical stage.
We therefore leverage their ordinal relationship as weak supervision to learn a continuous disease representation, allowing subjects with the same diagnosis to occupy different positions along the AD continuum. 
We formulate DCS learning based on two complementary structural assumptions.
At the population level, the latent patient-position distributions associated with the ordered clinical stages are expected to follow the same ordinal structure along the disease continuum, with 
\begin{equation}
    \mu_{\mathrm{y_{i}^{t}=1}} < \mu_{\mathrm{y_{i}^{t}=2}} < \mu_{\mathrm{y_{i}^{t}=3}},
\end{equation}
where \(\mu_{y_{i}^{t}}\) denotes the population-level latent anchor associated with diagnostic stage \(y\). 
At the individual level, we assume that the underlying disease-continuum position is non-decreasing over longitudinal follow-up \citep{jack2010hypothetical,jack2013tracking,schaap2024timing}. 
Specifically, for two observations of the same subject \(i\) acquired at \(t_a\) and \(t_b\), where \(t_b>t_a\), we assume
$s_i^{t_b}\geq s_i^{t_a}$.
This assumption characterizes the macroscopic progression of the underlying disease continuum rather than requiring individual neuroimaging measurements to exhibit monotonic changes \citep{aksman2019modeling,jack2013biomarker}. 
In practice, we incorporate this assumption as a soft longitudinal constraint to accommodate measurement variability and heterogeneous progression patterns.

\subsection{Framework Overview}
To estimate the DCS under the above population-level and longitudinal assumptions, we propose the Disease Continuum Positioning (DCP) framework, as illustrated in Figure~\ref{fig:framework}. 
Given a neuroimaging observation \(X_i^t\), DCP employs a ResNet-based encoder to map the input into a structured probabilistic latent representation
\begin{equation}
    z_i^t=\left[\tilde{s}_i^t,\,z_{r,i}^t\right]^\top \in \mathbb{R}^{d},
\end{equation}
where \(\tilde{s}_i^t\in\mathbb{R}\) is a one-dimensional disease-order-aware latent variable from which the non-negative DCS \(s_i^t\) is derived, and \(z_{r,i}^t\in\mathbb{R}^{d-1}\) denotes the residual latent variables that retain complementary imaging information.

DCP learns the disease-order-aware latent variable through two complementary forms of supervision. 
Diagnosis-anchored probabilistic priors and ordinal constraints establish population-level disease ordering, while follow-up observations of the same subject impose longitudinal constraints on within-subject progression over time. 
Meanwhile, the residual latent variables are regularized toward a standard Gaussian prior, together with a representation-preserving regularization to retain complementary imaging information. 
These components jointly enable DCP to estimate continuous patient positions that reflect both population-level disease ordering and individual longitudinal progression.

\subsection{Structured Probabilistic Representation}
DCP models the disease-order-aware variable \(\tilde{s}_i^t\) and the residual latent variables \(z_{r,i}^t\) probabilistically, with distinct prior distributions imposed according to their respective roles. The former is structured by diagnosis-conditioned priors to establish the disease-continuum axis, whereas the latter is regularized to preserve a compact residual latent space.

\subsubsection{Probabilistic Latent Encoding}
Given a neuroimaging observation \(X_i^t\), an encoder \(E_{\phi}\) parameterizes the approximate posterior distributions of the two latent components as
\begin{equation}
q_{\phi}\!\left(\tilde{s}_i^t \mid X_i^t\right)
=
\mathcal{N}\!\left(
\mu_{s,i}^t,\,
(\sigma_{s,i}^t)^2
\right),
\end{equation}
and
\begin{equation}
q_{\phi}\!\left(z_{r,i}^t \mid X_i^t\right)
=
\mathcal{N}\!\left(
\mu_{r,i}^t,\,
\operatorname{diag}\!\left((\sigma_{r,i}^t)^2\right)
\right),
\end{equation}
where \(\mu_{s,i}^t,\sigma_{s,i}^t\in\mathbb{R}\) and
\(\mu_{r,i}^t,\sigma_{r,i}^t\in\mathbb{R}^{d-1}\). 
Accordingly, the encoder predicts the mean and log-variance vectors
\begin{equation}
\mu_i^t
=
\left[
\mu_{s,i}^t,\,
\mu_{r,i}^t
\right]^\top,
\quad
\log(\sigma_i^t)^2
=
\left[
\log(\sigma_{s,i}^t)^2,\,
\log(\sigma_{r,i}^t)^2
\right]^\top.
\end{equation}
Latent samples are obtained through reparameterization:
\begin{equation}
\tilde{s}_i^t
=
\mu_{s,i}^t
+
\sigma_{s,i}^t\epsilon_{s,i}^t,
\qquad
\epsilon_{s,i}^t\sim\mathcal{N}(0,1),
\end{equation}
\begin{equation}
z_{r,i}^t
=
\mu_{r,i}^t
+
\sigma_{r,i}^t\odot\epsilon_{r,i}^t,
\qquad
\epsilon_{r,i}^t\sim\mathcal{N}(0,I_{d-1}),
\end{equation}
where \(\odot\) denotes element-wise multiplication, and
\(I_{d-1} \in \mathbb{R}^{(d-1)\times(d-1)}\) is an identity matrix. This formulation enables stochastic latent sampling and end-to-end optimization \citep{kingma2013auto}.

\subsubsection{Diagnosis-Anchored Priors}

To impose population-level disease ordering on the disease-order-aware latent variable, we define a diagnosis-conditioned Gaussian prior as
\begin{equation}
p\!\left(\tilde{s}_i^t \mid y_i^t\right)
=
\mathcal{N}\!\left(
\mu_{y_i^t},
\sigma_{pr}^2
\right),
\end{equation}
where \(\mu_y\) denotes the prior mean associated with diagnostic stage \(y\), and \(\sigma_{pr}=1\) denotes the shared prior standard deviation. 
The prior means are predefined according to the ordinal relationship among diagnostic stages:
\begin{equation}
y_{i_{1}}^{t_{1}} < y_{i_{2}}^{t_{2}}
\quad\Longrightarrow\quad
\mu_{y_{i_{1}}^{t_{1}}} < \mu_{y_{i_{2}}^{t_{2}}}.
\end{equation}
Thus, the diagnosis-conditioned priors provide ordered population-level anchors along the latent disease-continuum axis while allowing individual observations within the same diagnostic stage to occupy different latent positions. 
The corresponding KL regularization is
\begin{equation}
\mathcal{L}_{\mathrm{KL}}^{s}
=
D_{\mathrm{KL}}
\left(
q_{\phi}\!\left(\tilde{s}_i^t\mid X_i^t\right)
\,\Vert\,
p\!\left(\tilde{s}_i^t\mid y_i^t\right)
\right).
\end{equation}

For the residual latent variables, we adopt an isotropic Gaussian prior
\begin{equation}
p\!\left(z_{r,i}^t\right)
=
\mathcal{N}\!\left(
0,I_{d-1}
\right),
\end{equation}
with the corresponding KL regularization
\begin{equation}
\mathcal{L}_{\mathrm{KL}}^{r}
=
D_{\mathrm{KL}}
\left(
q_{\phi}\!\left(z_{r,i}^t\mid X_i^t\right)
\,\Vert\,
p\!\left(z_{r,i}^t\right)
\right).
\end{equation}
The overall latent regularization is defined as
\begin{equation}
\mathcal{L}_{\mathrm{KL}}
=
\mathcal{L}_{\mathrm{KL}}^{s}
+
\mathcal{L}_{\mathrm{KL}}^{r}.
\end{equation}
Here, \(\mathcal{L}_{\mathrm{KL}}^{s}\) structures the disease-order-aware latent variable according to diagnosis-conditioned population priors, whereas \(\mathcal{L}_{\mathrm{KL}}^{r}\) regularizes the residual latent variables without imposing explicit disease independence.

\subsection{Disease-Order-Aware Latent Structuring}
The diagnosis-conditioned priors introduced above provide distribution-level anchors for organizing the latent disease-continuum axis. 
However, these probabilistic priors alone do not explicitly enforce the relative ordering of latent positions between individual subjects. 
We therefore introduce a pairwise ordinal ranking constraint to further establish subject-level ordering across clinical stages.

For two subjects \(i\) and \(j\), with diagnostic stages \(y_i\) and \(y_j\) and corresponding disease-order-aware latent variables \(\tilde{s}_i\) and \(\tilde{s}_j\), the subject at the more advanced diagnostic stage is expected to occupy a higher latent position. 
We formulate this constraint using a margin-based ordinal ranking loss:
\begin{equation}
\begin{aligned}
\mathcal{L}_{\mathrm{rank}}
=
\mathbb{E}_{i,j}
\Big[
\max\Big(
0,\,
\gamma \left|y_i-y_j\right| -
\left(\tilde{s}_i-\tilde{s}_j\right)
\operatorname{sgn}\left(y_i-y_j\right)
\Big)
\Big],
\end{aligned}
\end{equation}
where \(\gamma>0\) is a margin hyperparameter controlling the minimum separation between latent positions across diagnostic stages. 
The term \(\left|y_i-y_j\right|\) scales the required margin according to the ordinal distance between the corresponding stages. 
Note that \(\mathcal{L}_{\mathrm{rank}}\) is evaluated using baseline observations only. Together with the diagnosis-conditioned priors, this ranking constraint encourages the latent disease-continuum axis to preserve both population-level stage ordering and subject-level ordinal relationships.

\subsection{Longitudinal Disease Progression Modeling}
While the disease-order-aware latent structuring establishes relative ordering across subjects, it does not explicitly model how an individual subject progresses along the disease continuum over time. To incorporate longitudinal disease evolution, DCP introduces a temporal monotonicity constraint over repeated observations from the same subject. Specifically, for subject \(i\), consider two observations acquired at \(t_a\) and \(t_b\), where \(t_b>t_a\). Following the progressive nature of AD, we assume that the underlying disease-continuum position is non-decreasing over time:
\begin{equation}
\tilde{s}_i^{t_b}
\geq
\tilde{s}_i^{t_a}.
\end{equation}

Rather than enforcing this assumption as a hard constraint, we introduce a soft monotonicity penalty on longitudinal pairs that violate the expected temporal ordering. To account for differences in follow-up duration, we further define a logarithmically scaled time-span weight:
\begin{equation}
\omega(t_a,t_b)
=
1+\log\left(1+\left|t_b-t_a\right|\right).
\end{equation}
Longitudinal pairs separated by a longer follow-up interval therefore receive a larger penalty when the inferred disease position violates the expected progression direction, while the logarithmic scaling prevents excessively large weights for long intervals. The longitudinal monotonicity loss is defined as
\begin{equation}
\mathcal{L}_{\mathrm{mono}}
=
\mathbb{E}_{i,t_a<t_b}
\left[
\omega(t_a,t_b)
\max
\left(
0,\,
-\left(
\tilde{s}_i^{t_b}
-
\tilde{s}_i^{t_a}
\right)
\right)
\right].
\end{equation}
This soft constraint incorporates within-subject temporal ordering into the learned disease continuum while allowing flexibility for measurement variability and heterogeneous longitudinal trajectories.

\subsection{Overall Learning Objective}
DCP is trained end-to-end by jointly optimizing the probabilistic latent regularization, ordinal ranking, longitudinal monotonicity, and representation-preserving objectives. To encourage the structured latent representation to retain complementary information from the input image, we introduce an auxiliary reconstruction regularization. Specifically, the latent representation \(z_i^t\) is projected to the spatial feature dimension, reshaped into a multi-dimensional feature tensor, and progressively mapped back to the image space through a deconvolutional decoder \(G_{\psi}\): $\hat{X}_i^t=
G_{\psi}\!\left(z_i^t\right).$
The reconstruction regularization is defined using the mean squared error (MSE): $\mathcal{L}_{\mathrm{rec}}=\left\|X_i^t-\hat{X}_i^t\right\|_2^2.$
This auxiliary objective encourages \(z_i^t\) to preserve imaging information while learning the structured disease-continuum representation.
The overall learning objective is formulated as
\begin{equation}
\begin{aligned}
\mathcal{L}_{\mathrm{total}}
={}
\lambda_{\mathrm{KL}}^{s}\mathcal{L}_{\mathrm{KL}}^{s}
+\lambda_{\mathrm{KL}}^{r}\mathcal{L}_{\mathrm{KL}}^{r}
+\lambda_{\mathrm{rank}}\mathcal{L}_{\mathrm{rank}} 
+
\lambda_{\mathrm{mono}}\mathcal{L}_{\mathrm{mono}}
+\lambda_{\mathrm{rec}}\mathcal{L}_{\mathrm{rec}},
\end{aligned}
\end{equation}
where the \(\lambda \geq 0\) coefficients control the relative contributions of the corresponding objectives.

\section{Experimental Results}
\subsection{Dataset and Data Preprocessing}
We utilized diffusion-weighted MRI (dMRI) data from the publicly available Alzheimer's Disease Neuroimaging Initiative (ADNI) database \citep{petersen2010alzheimer}. 
The dataset consisted of 695 participants, including 380 cognitively normal (CN) subjects (mean age = 73.6 ± 7.2 years, 211 female), 222 subjects with mild cognitive impairment (MCI) (mean age = 74.8 ± 8.2 years, 89 female), and 93 patients with Alzheimer's disease (AD) (mean age = 75.8 ± 8.7 years, 39 female). 
Each participant underwent between 1 and 13 longitudinal imaging visits, resulting in a total of 2036 dMRI scans.

The dMRI data were preprocessed using the FMRIB Software Library (FSL)\citep{jenkinson2012fsl}. 
Briefly, the preprocessing pipeline\citep{zhan2015comparison} included brain extraction, motion and eddy-current correction, susceptibility-induced distortion correction, and diffusion tensor fitting. 
Four diffusion tensor-derived parametric maps\citep{zhan2012spatial}, including fractional anisotropy (FA), mean diffusivity (MD), radial diffusivity (RD), and axial diffusivity (AD), were reconstructed and used as the input of our model. 
The effects of demographic covariates, including age, sex, and years of education, were removed using general linear regression \cite{dukart2011age}. 
The resulting residual features were used for all subsequent analyses.

\subsection{Implementation Details}
Prior to model training, we controlled for potential demographic confounders by regressing out the effects of age, sex, and years of education from the diffusion-derived scalar maps using a voxel-wise ordinary least squares (OLS) general linear model \citep{dukart2013relationship,dukart2011age}. 
Following established neuroimaging practices~\citep{dukart2011age}, the regression coefficients were estimated exclusively from cognitively normal (CN) subjects to reduce the risk of removing disease-related variation that may covary with demographic effects. 
The estimated coefficients were subsequently applied to all observations, and the resulting residualized FA, MD, RD, and AD maps were used as the four-channel inputs to DCP.

DCP was implemented using a 3D ResNet-18 encoder that parameterizes a \(d\)-dimensional probabilistic latent representation. 
We set \(d=128\), consisting of the one-dimensional disease-order-aware latent variable \(\tilde{s}_i^t\) and 127 residual latent variables \(z_{r,i}^t\). 
Following latent inference, the reported DCS was obtained by applying a Softplus transformation:
\begin{equation}
s_i^t
=
\operatorname{Softplus}\!\left(\tilde{s}_i^t\right).
\end{equation}
This transformation constrains the reported DCS to be non-negative while preserving the disease ordering learned in the latent space, since Softplus is strictly monotonic.

During training, data augmentation was performed on the fly using random axial rotations within \([-5^{\circ},5^{\circ}]\), channel-wise intensity perturbations of up to \(3\%\) of the maximum intensity, and Gaussian noise scaled according to the variance of each channel. 
All input maps were resampled to a spatial dimension of \(91\times109\times91\) using trilinear interpolation. 
The model was trained end-to-end for 80 epochs using the AdamW optimizer with an initial learning rate of \(1.5\times10^{-4}\), which was decayed to \(1\times10^{-6}\) using a cosine annealing schedule, and a weight decay of \(0.05\). 
Each mini-batch contained up to 96 longitudinal observations. The longitudinal loss weight \(\lambda_{\mathrm{mono}}\) was set to zero during the first five epochs and subsequently linearly increased to its target value, progressively introducing the longitudinal constraint during training. 
Hyperparameters associated with the learning objective, including the loss-balancing weights and ordinal ranking margin, were automatically optimized using Optuna \citep{akiba2019optuna}.

Model performance was evaluated using subject-level five-fold cross-validation. 
All longitudinal observations from the same subject were assigned to the same fold, ensuring complete subject-level separation between the training and test sets. 
All experiments were implemented in PyTorch 2.5.1 with CUDA 12.1 and conducted on 2 NVIDIA A100 GPUs.
The source code for our DCP is publicly available at 
\url{https://github.com/YingyingZhang85/DCP.git}

\begin{table*}[t]      
\centering
\caption{
Quantitative comparison of DCP with baseline methods. Spearman's $\rho$ measures the association between the derived DCS and MMSE, ANOVA $F$ evaluates population-level 
variation across CN, MCI, and AD, and Stage C and Long.\ C quantify concordance 
with diagnostic-stage ordering and within-subject longitudinal ordering, 
respectively. Avg.\ C denotes the average of Stage C and Long.\ C. 
Best and second best performance are highlighted in \textcolor{red}{red} and \textcolor{blue}{blue}. 
}
\label{tab:comparative_results}
\resizebox{\textwidth}{!}{
\begin{tabular}{lccccc}
\toprule
Methods 
& Spearman $\rho$ (vs.\ MMSE)
& ANOVA $F$ $\uparrow$
& Stage C $\uparrow$
& Long.\ C $\uparrow$
& Avg.\ C $\uparrow$ \\
\midrule

SuStaIn~\citep{young2018sustain}
& \textcolor{blue}{-0.4138 $\pm$ 0.0729}
& \textcolor{blue}{35.97 $\pm$ 11.56}
& \textcolor{blue}{0.7194 $\pm$ 0.0417}
& \textcolor{blue}{0.7879 $\pm$ 0.0116}
& \textcolor{blue}{0.7536 $\pm$ 0.0200} \\

DLMRI~\citep{feng2022deep}
& -0.1994 $\pm$ 0.0890
& 22.47 $\pm$ 13.89
& 0.6590 $\pm$ 0.0354
& 0.3840 $\pm$ 0.0554
& 0.5215 $\pm$ 0.0360 \\

LSSL~\citep{ouyang2022longitudinal}
& -0.2266 $\pm$ 0.0756
& 7.7884 $\pm$ 4.3366
& 0.6092 $\pm$ 0.0489
& 0.7612 $\pm$ 0.0141
& 0.6855 $\pm$ 0.0190 \\

MedicalNet~\citep{natureaging2026}
& -0.3396 $\pm$ 0.1068
& 14.7468 $\pm$ 8.0318
& 0.7003 $\pm$ 0.0581
& -- 
& -- \\

\textbf{DCP (Ours)}
& \textcolor{red}{-0.5481 $\pm$ 0.0748}
& \textcolor{red}{40.76 $\pm$ 8.68}
& \textcolor{red}{0.7794 $\pm$ 0.0063}
& \textcolor{red}{0.8367 $\pm$ 0.0187}
& \textcolor{red}{0.8076 $\pm$ 0.0135} \\

\bottomrule
\end{tabular}}
\end{table*}

\subsection{Baseline Methods}
We compared DCP with four representative methods covering complementary paradigms of disease progression modeling and continuous severity estimation: Subtype and Stage Inference (SuStaIn)~\citep{young2018sustain}, Deep Learning MRI (DLMRI)~\citep{feng2022deep}, Longitudinal Self-Supervised Learning (LSSL)~\citep{ouyang2022longitudinal}, and MedicalNet~\citep{natureaging2026}. \\
\textbf{Subtype and Stage Inference (SuStaIn)}~\cite{young2018sustain} is a data-driven disease progression model that jointly identifies heterogeneous disease subtypes and estimates disease stages from the sequential occurrence of biomarker abnormalities. By modeling multiple possible progression patterns, SuStaIn provides an unsupervised framework for positioning individuals along subtype-specific disease trajectories. \\
\textbf{Deep Learning MRI (DLMRI)}~\cite{feng2022deep} is a supervised deep learning framework for estimating continuous AD severity from neuroimaging data. The model learns an imaging-derived severity score under diagnostic supervision and subsequently adjusts the predicted score for demographic covariates, providing a continuous representation of disease severity beyond categorical diagnosis. \\
\textbf{Longitudinal Self-Supervised Learning (LSSL)}~\cite{ouyang2022longitudinal} is a longitudinal representation learning framework that exploits repeated observations from the same subject to learn disease-related changes over time. The method incorporates longitudinal self-supervision to structure the latent representation and separate disease-related severity variation from normal aging effects. \\
\textbf{MedicalNet}~\cite{natureaging2026} is a deep learning framework for predicting continuous clinical outcomes from baseline neuroimaging observations. Its original architecture employs multi-task learning, combining continuous outcome prediction with auxiliary tissue-segmentation supervision to learn clinically relevant imaging representations. \\
\noindent\textbf{Baseline Adaptations.}
To apply the baseline methods to our DTI cohort while preserving their original modeling paradigms, we introduced only the adaptations necessary for compatibility with the available data. 
The official implementations provided by the respective studies were used whenever available. 
For SuStaIn \cite{young2018sustain}, which operates on regional biomarker measurements rather than voxel-wise images, we used six FreeSurfer-derived volumetric measures available in ADNI-MERGE: Ventricles, Hippocampus, WholeBrain, Entorhinal, Fusiform, and MidTemp. 
Each regional measure was represented by three z-score events, yielding an 18-event progression sequence. 
The inferred SuStaIn stage was subsequently used as its disease progression estimate.
For all the other 3 baselines, FA was used as the imaging input in accordance with their single-modality input configurations. 
DLMRI \cite{feng2022deep} was trained using the original FA maps, following its original strategy of performing demographic adjustment on the predicted severity scores. 
Specifically, the effects of age, sex, and education were regressed out from the predicted scores using linear regression. 
Longitudinal observations were treated as independent inputs, consistent with the original model formulation.
LSSL \cite{ouyang2022longitudinal} was trained using the covariate-residualized FA maps and retained its original longitudinal learning strategy based on repeated observations from the same subject. 
MedicalNet \cite{natureaging2026} was also trained using the covariate-residualized FA maps. 
Because tissue-segmentation targets required by its auxiliary segmentation branch were unavailable for the DTI inputs, this branch was disabled, while the continuous outcome prediction branch was retained. 
Consistent with its original prognostic setting, MedicalNet was trained using baseline observations only.
We implemented and adapted all baseline methods based on the official code released by the respective authors (i.e.,SuStaIn\footnote{\url{https://github.com/ucl-pond/pySuStaIn}}, DLMRI \footnote{\url{https://github.com/fengcls/Deep-learning-MRI-AD-prediction}}, LSSL\footnote{\url{https://github.com/ouyangjiahong/longitudinal-direction-disentangle}}, and MedicalNet \footnote{\url{https://github.com/darenma/MultitaskCognition}}).

\subsection{Evaluation Metrics}
We evaluated the learned DCS from three complementary perspectives: its association with cognitive function, its population-level variation across the AD continuum, and its consistency with longitudinal disease ordering.
First, we assessed the association between DCS and the Mini-Mental State Examination (MMSE) \citep{gallegos202245}, using Spearman's rank correlation coefficient (\(\rho\)).
Second, we examined the population-level variation of DCS across the AD continuum using one-way analysis of variance (ANOVA) among CN, MCI, and AD. 
The \(F\)-statistic quantifies between-stage variation relative to within-stage variability, with a larger value indicating a stronger association between DCS and the clinically defined disease continuum.
Finally, motivated by concordance-based evaluation of AD staging and progression~\citep{li2019deep,wang2024learning}, we used two concordance-based metrics to assess disease ordering. Stage C measures the consistency between DCS ordering and ordinal diagnostic stages across subjects, whereas Long. C measures the consistency between DCS ordering and the temporal ordering of repeated observations within subjects. Higher values indicate greater ordering consistency.

\begin{table*}[htbp]
\centering
\caption{
Ablation study of DCP with respect to objective components and DTI input measures.
The objective component ablation evaluates the contributions of the ordinal ranking
loss $\mathcal{L}_{\mathrm{rank}}$, longitudinal monotonicity loss
$\mathcal{L}_{\mathrm{mono}}$, and probabilistic latent regularization
$\mathcal{L}_{\mathrm{KL}}$. The DTI modality ablation evaluates DCP using each
DTI-derived measure individually (FA, MD, AD, or RD), compared with the full model
using all four measures. 
}
\label{tab:ablation}
\resizebox{\textwidth}{!}{
\begin{tabular}{lccccc}
\toprule
Setting
& Spearman $\rho$ (vs.\ MMSE)
& ANOVA $F$ $\uparrow$
& Stage C $\uparrow$
& Long.\ C $\uparrow$
& Avg.\ C $\uparrow$ \\
\midrule

\textbf{DCP (Full)}
& \textcolor{black}{-0.5481 $\pm$ 0.0748}
& \textcolor{black}{40.76 $\pm$ 8.68}
& \textcolor{black}{0.7794 $\pm$ 0.0063}
& \textcolor{black}{0.8367 $\pm$ 0.0187}
& \textcolor{black}{0.8076 $\pm$ 0.0135}\\

\midrule
\multicolumn{6}{l}{\emph{Objective Component Ablation}} \\

\quad w/o $\mathcal{L}_{\mathrm{rank}}$
& -0.4235 $\pm$ 0.0823
& 28.33 $\pm$ 9.04
& 0.7330 $\pm$ 0.0267
& 0.8686 $\pm$ 0.0112
& 0.7990 $\pm$ 0.0131 \\

\quad w/o $\mathcal{L}_{\mathrm{mono}}$
& -0.5084 $\pm$ 0.0307
& 41.42 $\pm$ 2.70
& 0.8071 $\pm$ 0.0058
& 0.6713 $\pm$ 0.0109
& 0.7377 $\pm$ 0.0096 \\

\quad w/o $\mathcal{L}_{\mathrm{KL}}$
& -0.4239 $\pm$ 0.0612
& 30.15 $\pm$ 7.42
& 0.7673 $\pm$ 0.0240
& 0.7605 $\pm$ 0.0451
& 0.7612 $\pm$ 0.0292 \\

\midrule
\multicolumn{6}{l}{\emph{DTI Measure Ablation}} \\

\quad FA only
& -0.4366 $\pm$ 0.0928
& 38.23 $\pm$ 6.95
& 0.7720 $\pm$ 0.0074
& 0.7381 $\pm$ 0.0119
& 0.7538 $\pm$ 0.0095 \\

\quad MD only
& -0.4347 $\pm$ 0.0897
& 26.50 $\pm$ 5.42
& 0.7606 $\pm$ 0.0254
& 0.7777 $\pm$ 0.0288
& 0.7666 $\pm$ 0.0281 \\

\quad AD only
& -0.4685 $\pm$ 0.0978
& 22.63 $\pm$ 2.43
& 0.7645 $\pm$ 0.0271
& 0.7527 $\pm$ 0.0159
& 0.7557 $\pm$ 0.0231 \\

\quad RD only
& -0.4676 $\pm$ 0.1102
& 33.15 $\pm$ 3.76
& 0.7732 $\pm$ 0.0202
& 0.7454 $\pm$ 0.0528
& 0.7593 $\pm$ 0.0365 \\

\bottomrule
\end{tabular}}
\end{table*}

\subsection{Comparative Experiments}
Table~\ref{tab:comparative_results} presents the quantitative comparison of Disease Continuum Positioning (DCP) with the four baseline methods. 
Overall, DCP achieves the best performance across all evaluated metrics. 
The DCP-derived Disease Continuum Score (DCS) shows the strongest association with cognitive function, achieving a Spearman correlation of \(\rho=-0.5481\) with MMSE (\(p=2.16\times10^{-30}\)). 
The negative association indicates that a higher DCS, corresponding to a more advanced position along the AD continuum, is associated with greater cognitive impairment. 
DCS also exhibits the strongest population-level variation across the clinically defined AD continuum, achieving the highest ANOVA \(F\)-statistic of \(40.76\), compared with \(35.97\) for the most competitive baseline, SuStaIn~\citep{young2018sustain}.

DCP further demonstrates consistent disease ordering at both the population and longitudinal levels. 
It achieves the highest Stage C of \(0.7794\), outperforming SuStaIn (\(0.7194\)), MedicalNet (\(0.7003\)), DLMRI (\(0.6590\)), and LSSL (\(0.6092\)). 
For longitudinal ordering, DCP achieves the highest Long.\ C of \(0.8367\), exceeding SuStaIn (\(0.7879\)) and LSSL (\(0.7612\)), despite the latter being specifically designed for longitudinal representation learning. 
Consequently, DCP achieves the highest Avg.\ C of \(0.8076\), demonstrating its ability to jointly preserve disease-stage ordering across subjects and temporal progression within subjects. 
Together, these results show that DCP learns a continuous disease representation that is consistently associated with cognitive function while capturing both population-level disease organization and within-subject longitudinal progression.

\subsection{Ablation Studies}
Table~\ref{tab:ablation} evaluates the contributions of the major objective components and DTI measures in DCP. 
For the objective component ablation, removing $\mathcal{L}_{\mathrm{rank}}$ decreases Stage C from $0.7794$ to $0.7330$ and weakens both the association with MMSE and the population-level variation across the AD continuum. 
Interestingly, Long.\ C increases to $0.8686$, suggesting that emphasizing longitudinal progression alone does not necessarily preserve disease-stage ordering across subjects. 
Conversely, removing $\mathcal{L}_{\mathrm{mono}}$ increases Stage C to $0.8071$ but substantially reduces Long.\ C from $0.8367$ to $0.6713$, confirming the critical role of the longitudinal constraint in preserving within-subject temporal ordering. 
Removing $\mathcal{L}_{\mathrm{KL}}$ leads to consistent degradation across all evaluated metrics, indicating that the structured probabilistic priors contribute to organizing the learned disease-continuum representation. 
Together, these results demonstrate the complementary roles of the three objectives: $\mathcal{L}_{\mathrm{rank}}$ promotes disease-stage ordering, $\mathcal{L}_{\mathrm{mono}}$ enforces longitudinal progression consistency, and $\mathcal{L}_{\mathrm{KL}}$ provides probabilistic structure to the latent continuum.

The DTI measure ablation further evaluates whether complementary diffusion information contributes to DCS estimation. 
Using all four DTI measures jointly yields the strongest association with MMSE ($\rho=-0.5481$), the highest Stage C ($0.7794$), and the highest Avg.\ C ($0.8076$) among the evaluated input configurations. 
Although individual DTI measures can achieve competitive performance on specific metrics, no single measure consistently performs best across the evaluated dimensions. 
These results indicate that FA, MD, AD, and RD provide complementary information for characterizing the AD continuum, supporting their joint integration in DCP.

\subsection{Disease Continuum Characterization}
\begin{figure*}[t]
  \centering
  \includegraphics[width=\linewidth]{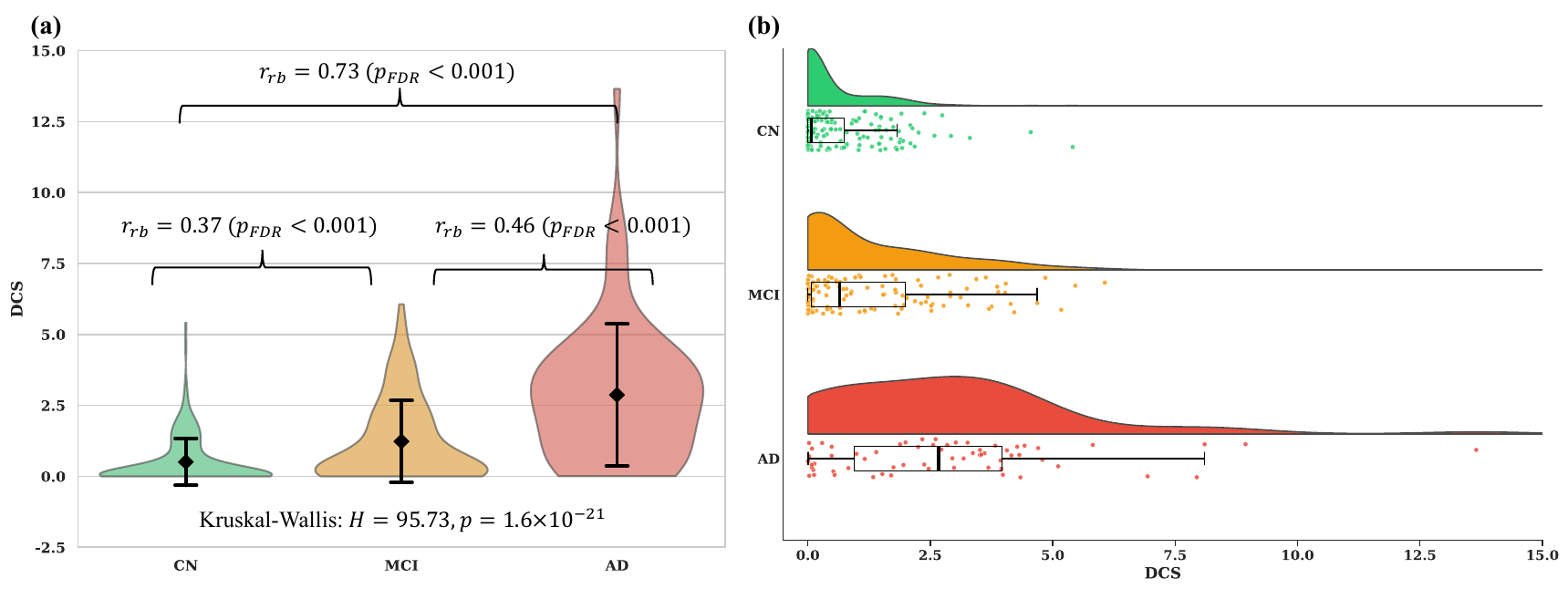}
  \caption{Distributional characterization of the Disease Continuum Score (DCS) across the clinically defined AD continuum.
(a) DCS distributions for CN, MCI, and AD. Group differences were assessed using the Kruskal--Wallis test followed by FDR-corrected pairwise Mann--Whitney $U$ tests; $r_{\mathrm{rb}}$ denotes the rank-biserial effect size.
(b) Raincloud plots showing individual DCS values and their distributions within each diagnostic group, illustrating both within-stage variability and overlap between adjacent diagnostic stages.
}
  \label{fig:47_disease_continuum}
\end{figure*}
We further characterized how the baseline Disease Continuum Score (DCS) is distributed across the clinically defined AD continuum. 
As shown in Figure \ref{fig:47_disease_continuum}(a), the DCS distributions exhibit a progressive shift from CN to MCI and AD. 
A Kruskal--Wallis test identified a significant overall difference among the three diagnostic groups (\(H=95.73\), \(p=1.6\times10^{-21}\)). 
Post-hoc pairwise Mann--Whitney \(U\) tests with false discovery rate (FDR) correction further showed significant differences for all pairwise comparisons (all \(p<0.001\)). 
The corresponding rank-biserial effect sizes (i.e., $r_{rb}$) increased with the clinical distance between groups, from CN versus MCI (\(r_{rb}=0.37\)) and MCI versus AD (\(r_{rb}=0.46\)) to CN versus AD (\(r_{rb}=0.73\)). 
Together with the progressive distributional shift, these results demonstrate that DCS exhibits systematic population-level organization along the clinically defined AD continuum, with the strongest distinction observed between CN and AD.

Beyond differences across diagnostic stages, we examined whether DCS preserves individual variation within each stage. 
The raincloud plots in Figure \ref{fig:47_disease_continuum}(b) reveal substantial within-stage variability in baseline DCS, together with overlapping distributions between adjacent diagnostic groups. 
Thus, subjects assigned to the same clinical diagnosis can occupy different positions along the learned continuum, while subjects in adjacent stages may exhibit similar DCS values. 
The coexistence of systematic between-stage shifts and within-stage variability supports the interpretation of DCS as a continuous measure of disease position rather than a direct encoding of categorical diagnostic labels.

\subsection{Longitudinal Progression Analysis}
We further investigated longitudinal changes in DCS by quantifying its time-normalized change between consecutive visits. 
For each subject, the DCS progression rate between two consecutive visits was calculated as
\begin{equation}
\Delta_{\mathrm{DCS},i}^{t_a,t_b}
=
\frac{
\mathrm{DCS}_i^{t_b}-\mathrm{DCS}_i^{t_a}
}{
t_b-t_a
},
\qquad t_b>t_a,
\end{equation}
where \(t_b-t_a\) denotes the inter-visit interval in days. 
The resulting progression rates were then grouped according to the subjects' baseline diagnoses, and the group-wise mean rates were calculated for CN, MCI, and AD.
The mean progression rates were
\(2.79\times10^{-4}\) DCS/day (or \(0.102\) DCS/year) for CN,
\(3.75\times10^{-4}\) DCS/day (or \(0.137\) DCS/year) for MCI, and
\(2.43\times10^{-3}\) DCS/day (or \(0.888\) DCS/year) for AD.
All three groups exhibited positive average longitudinal changes, with relatively modest rates in CN and MCI and a substantially greater rate in AD \citep{gong2026early,nowrangi2013longitudinal,jack2000rates}.
These results demonstrate that DCS captures time-dependent changes along the disease continuum and suggest that the group-level rate of change varies across baseline disease stages.

\subsection{Clinical Validation}
\begin{figure*}[h]
  \centering
  \includegraphics[width=\linewidth]{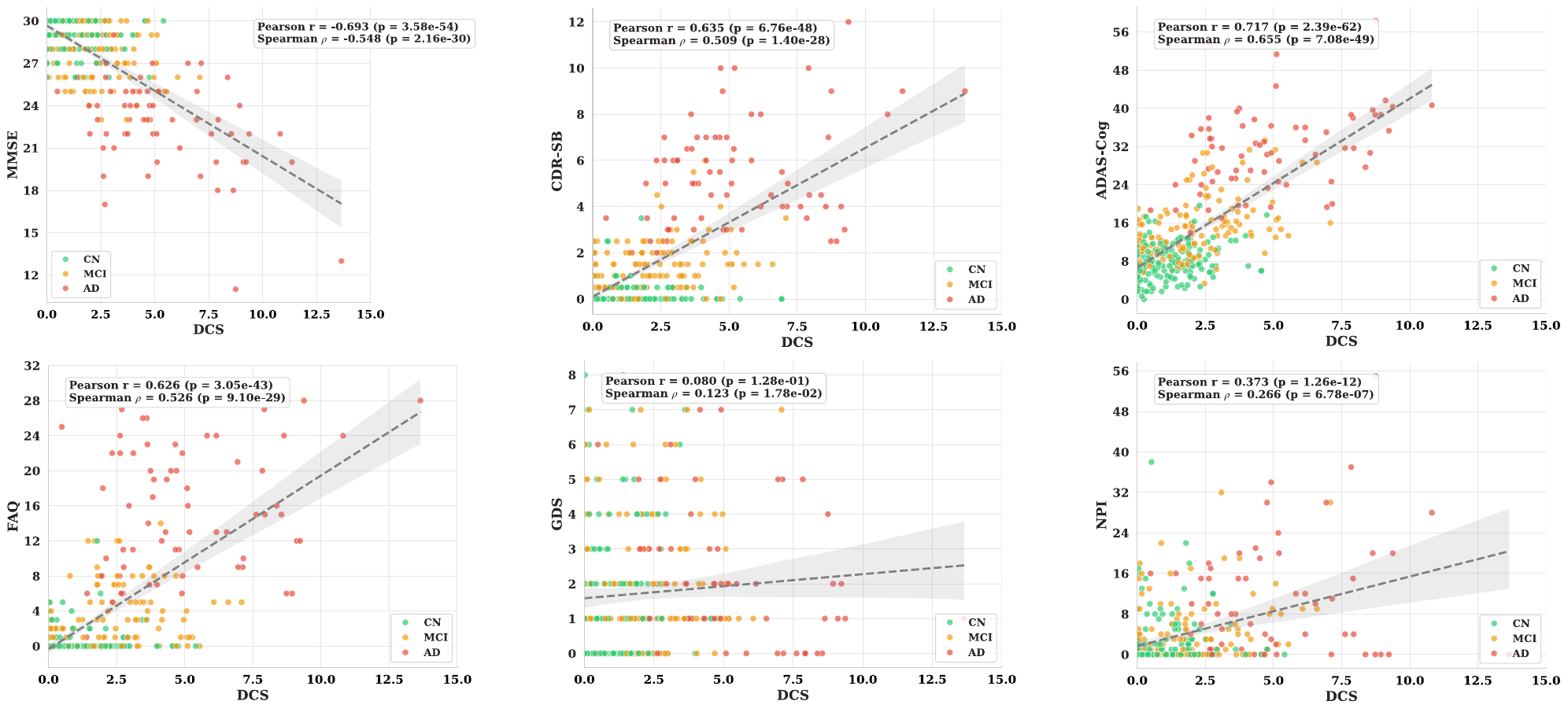}
  \caption{Clinical validation of the Disease Continuum Score (DCS) through its associations with established clinical assessments. Scatter plots show the relationships between DCS and MMSE, CDR-SB, ADAS-Cog-13, FAQ, GDS, and NPI, with subjects colored according to diagnostic group (CN, MCI, and AD). Pearson's correlation coefficient ($r$) and Spearman's rank correlation coefficient ($\rho$), together with their corresponding $p$-values, are reported for each assessment.}
  \label{fig:49_phenotypes_corr}
\end{figure*}
We further evaluated the clinical relevance of DCS by examining its associations with six established clinical assessments spanning cognitive function, dementia severity, functional impairment, and neuropsychiatric symptoms.
Pearson's correlation coefficient ($r$) and Spearman's rank correlation coefficient ($\rho$) were used to characterize linear and monotonic associations, respectively.
Figure~\ref{fig:49_phenotypes_corr} presents the relationships between DCS and the six clinical measures, with the corresponding correlation coefficients and statistical significance reported in each panel.

DCS showed strong and consistent associations with measures of cognitive and functional impairment. 
Higher DCS was associated with lower MMSE \citep{gallegos202245} ($r=-0.693$, $\rho=-0.548$) and higher ADAS-Cog-13 \citep{mohs1997development} ($r=0.717$, $\rho=0.655$), CDR-SB \citep{hughes1982new} ($r=0.635$, $\rho=0.509$), and FAQ \citep{pfeffer1982measurement} ($r=0.626$, $\rho=0.526$), with all associations being highly significant ($p<0.001$).
Importantly, the directions of these associations were clinically consistent: a more advanced DCS position was associated with poorer global cognition, greater cognitive impairment and dementia severity, and greater impairment in daily functioning.
Among these measures, ADAS-Cog-13 exhibited the strongest association with DCS, further supporting the sensitivity of the learned continuum to variation in cognitive impairment.

In contrast, DCS showed substantially weaker associations with neuropsychiatric measures. 
The association with NPI \citep{cummings2020neuropsychiatric} was modest ($r=0.373$, $\rho=0.266$), although statistically significant, whereas the association with GDS \citep{lanctot2024association} was minimal ($r=0.080$, $p=0.128$; $\rho=0.123$, $p=0.0178$).
This differential association pattern suggests that DCS is more closely related to the cognitive and functional dimensions of the AD continuum than to depressive or broader neuropsychiatric symptom burden.

Notably, these clinical assessments are not directly used to define the
continuous DCS during inference; rather, DCS is derived from neuroimaging and its learned disease-continuum representation.
The observed associations therefore provide external clinical evidence that
the imaging-derived continuum captures variation that is meaningfully related to patients' cognitive and functional status.
From a biological perspective, these findings further suggest that the DTI-derived disease representation captures brain structural alterations that covary with clinically manifested disease burden, although DCS should not be interpreted as a direct measure of any specific neuropathological process.
Together, the results support DCS as a clinically meaningful continuous representation that complements categorical diagnosis by quantifying individual position along the AD continuum.

\subsection{Prediction of Future Clinical Conversion}
\begin{figure*}[h]
  \centering
  \includegraphics[width=\linewidth]{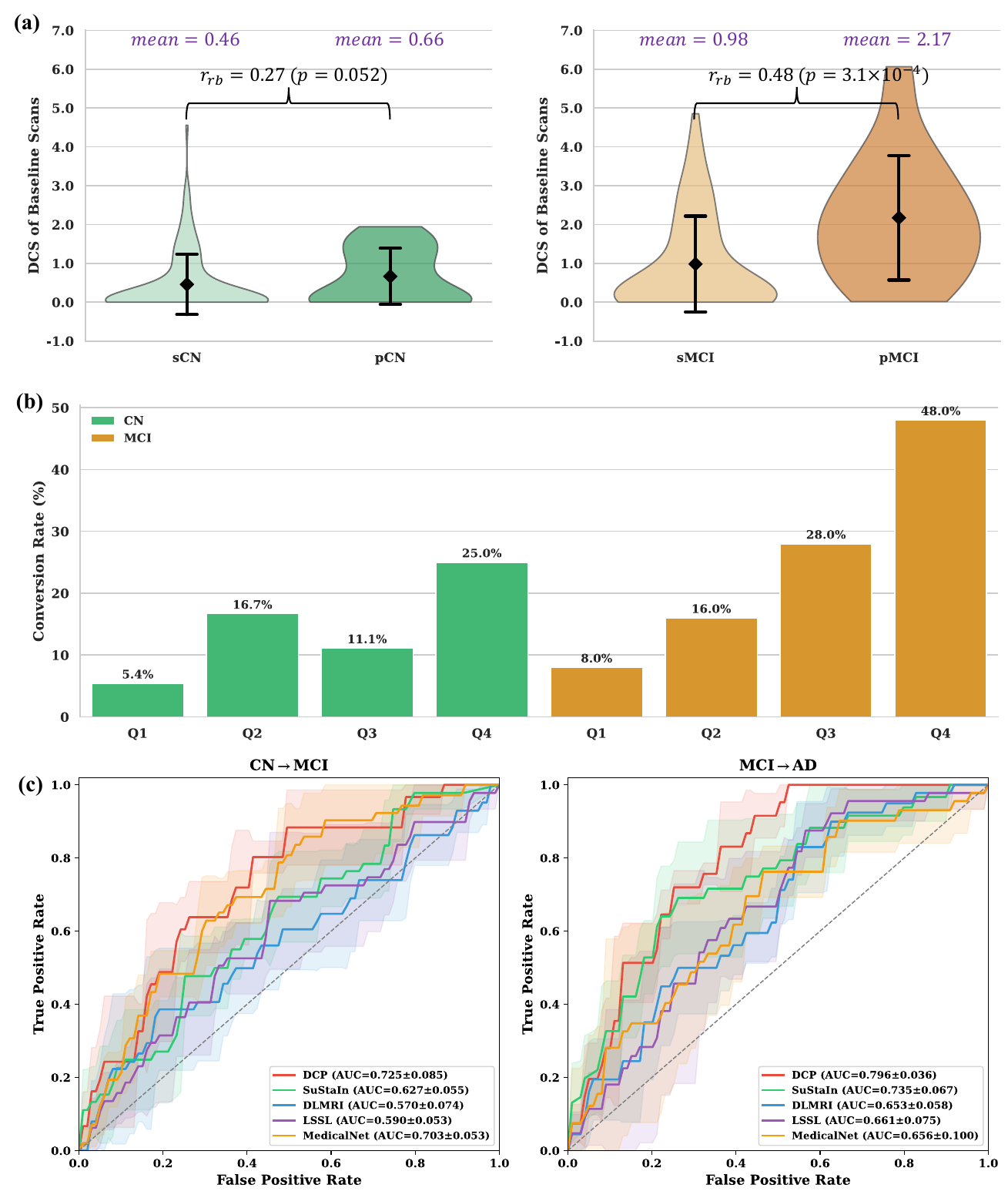}
  \caption{Prospective analysis of future clinical conversion using the Disease Continuum Score (DCS) derived from baseline scans.
(a) Baseline DCS distributions between stable and progressive subjects for CN$\rightarrow$MCI and MCI$\rightarrow$AD conversion. Group differences were assessed using the Mann--Whitney $U$ test, with $r_{\mathrm{rb}}$ denoting the rank-biserial effect size.
(b) Observed conversion rates after stratifying subjects into quartiles according to baseline DCS, from Q1 (lowest) to Q4 (highest), separately within the CN and MCI cohorts.
(c) Receiver operating characteristic (ROC) curves for CN$\rightarrow$MCI and MCI$\rightarrow$AD conversion. For each method, the disease score derived from the baseline scan was used as the sole predictor in logistic regression, with AUC reported as mean $\pm$ standard deviation across cross-validation folds.
Solid lines represent the mean ROC curves, shaded regions indicate variability across folds, and the diagonal dashed line represents chance-level discrimination.
}
\label{fig:410_conversion}
\end{figure*}
We further evaluated whether DCS derived from the baseline scan contains prospective information about future clinical conversion.
Subjects were stratified according to their baseline diagnosis and subsequent clinical outcome into stable CN (sCN), progressive CN (pCN; CN$\rightarrow$MCI), stable MCI (sMCI), and progressive MCI (pMCI; MCI$\rightarrow$AD). 
We first compared baseline DCS between stable and progressive subjects within each diagnostic group.

Among subjects with baseline MCI ($n=100$; 75 sMCI and 25 pMCI), pMCI subjects had a substantially higher baseline DCS than sMCI subjects (2.17 vs.\ 0.98).
The difference was supported by the Mann--Whitney $U$ test ($U=484$, $p=3.1\times10^{-4}$), with a rank-biserial effect size of $r_{rb}=0.48$, indicating that subjects who subsequently progressed to AD were already positioned further along the learned disease continuum at baseline.
A directionally consistent, although more modest, difference was observed in the CN cohort ($n=145$; 124 sCN and 21 pCN), where subjects who subsequently converted to MCI had a higher baseline DCS than stable CN subjects (0.66 vs.\ 0.46; $U=956$, $p=0.052$, $r_{rb}=0.27$).
The corresponding baseline DCS distributions are shown in Figure \ref{fig:410_conversion} (a). 

We next examined whether DCS derived from the baseline scan contains information about subsequent clinical progression.
Specifically, we trained a logistic regression model using DCS as the sole predictor of future conversion status.
The predicted conversion probabilities were used to construct receiver operating characteristic (ROC) curves, and the area under the ROC curve (AUC) was used to quantify the discrimination between stable and progressive subjects.
As shown in Figure \ref{fig:410_conversion}(c), DCP consistently achieved higher ROC performance than the baseline methods for both conversion tasks.
Quantitatively, as summarized in Table \ref{tab:conversion_auc}, DCP achieved an AUC of $0.725\pm0.085$ for CN$\rightarrow$MCI conversion and $0.796\pm0.036$ for MCI$\rightarrow$AD conversion.
For CN$\rightarrow$MCI, this exceeded SuStaIn ($0.627\pm0.055$), DLMRI ($0.570\pm0.074$), LSSL ($0.590\pm0.053$), and MedicalNet ($0.703\pm0.053$).
Similarly, for MCI$\rightarrow$AD, DCP achieved the highest AUC compared with SuStaIn ($0.735\pm0.067$), DLMRI ($0.653\pm0.058$), LSSL ($0.661\pm0.075$), and MedicalNet ($0.656\pm0.100$).
These results indicate that DCS derived from a single baseline scan contains prospective information about subsequent diagnostic progression, with stronger discrimination observed for MCI$\rightarrow$AD conversion.

Finally, we examined whether baseline DCS stratifies subjects according to future conversion risk. 
Within each baseline diagnostic group, subjects were divided into quartiles according to their baseline DCS, from Q1 (lowest) to Q4 (highest). 
For MCI, the observed conversion rate increased monotonically from $8.0\%$ in Q1 to $16.0\%$, $28.0\%$, and $48.0\%$ in Q2--Q4, respectively.
Thus, subjects in the highest DCS quartile exhibited a $6$-fold higher observed conversion rate than those in the lowest quartile. 
For CN, the corresponding conversion rates were $5.4\%$, $16.7\%$, $11.1\%$, and $25.0\%$ from Q1 to Q4.
Although the CN pattern was not strictly monotonic, subjects in the highest quartile showed an approximately $4.6$-fold higher observed conversion rate than those in the lowest quartile. 
These findings (see Figure \ref{fig:410_conversion}(b)) further support the prospective value of DCS: a more advanced baseline position along the learned disease continuum is associated with a greater likelihood of subsequent clinical progression, particularly among subjects with MCI.

\begin{table*}[htbp]
\centering
\caption{
Comparison of future clinical conversion prediction using disease scores derived 
from the baseline scan. For each method, the derived continuous score was used 
as the sole predictor in a logistic regression model to distinguish stable from 
progressive subjects. Performance was evaluated for CN$\rightarrow$MCI and 
MCI$\rightarrow$AD conversion using the area under the receiver operating 
characteristic curve (AUC), reported as mean $\pm$ standard deviation across 
cross-validation folds. The best performance is highlighted in bold.
}
\label{tab:conversion_auc}
\begin{tabular}{lcc}
\toprule
Method 
& AUC (CN$\rightarrow$MCI) $\uparrow$ 
& AUC (MCI$\rightarrow$AD) $\uparrow$ \\
\midrule

SuStaIn~\citep{young2018sustain}
& 0.6266 $\pm$ 0.0551
& 0.7345 $\pm$ 0.0674 \\

DLMRI~\citep{feng2022deep}
& 0.5695 $\pm$ 0.0736
& 0.6531 $\pm$ 0.0576 \\

LSSL~\citep{ouyang2022longitudinal}
& 0.5900 $\pm$ 0.0535
& 0.6615 $\pm$ 0.0745 \\

MedicalNet~\citep{natureaging2026}
& 0.7027 $\pm$ 0.0531
& 0.6564 $\pm$ 0.1000 \\

\textbf{DCP (Ours)}
& \textcolor{red}{0.7246 $\pm$ 0.0851}
& \textcolor{red}{0.7957 $\pm$ 0.0355} \\

\bottomrule
\end{tabular}
\end{table*}

\subsection{Uncertainty Analysis}
\begin{figure*}[h]
  \centering
  \includegraphics[width=\linewidth]{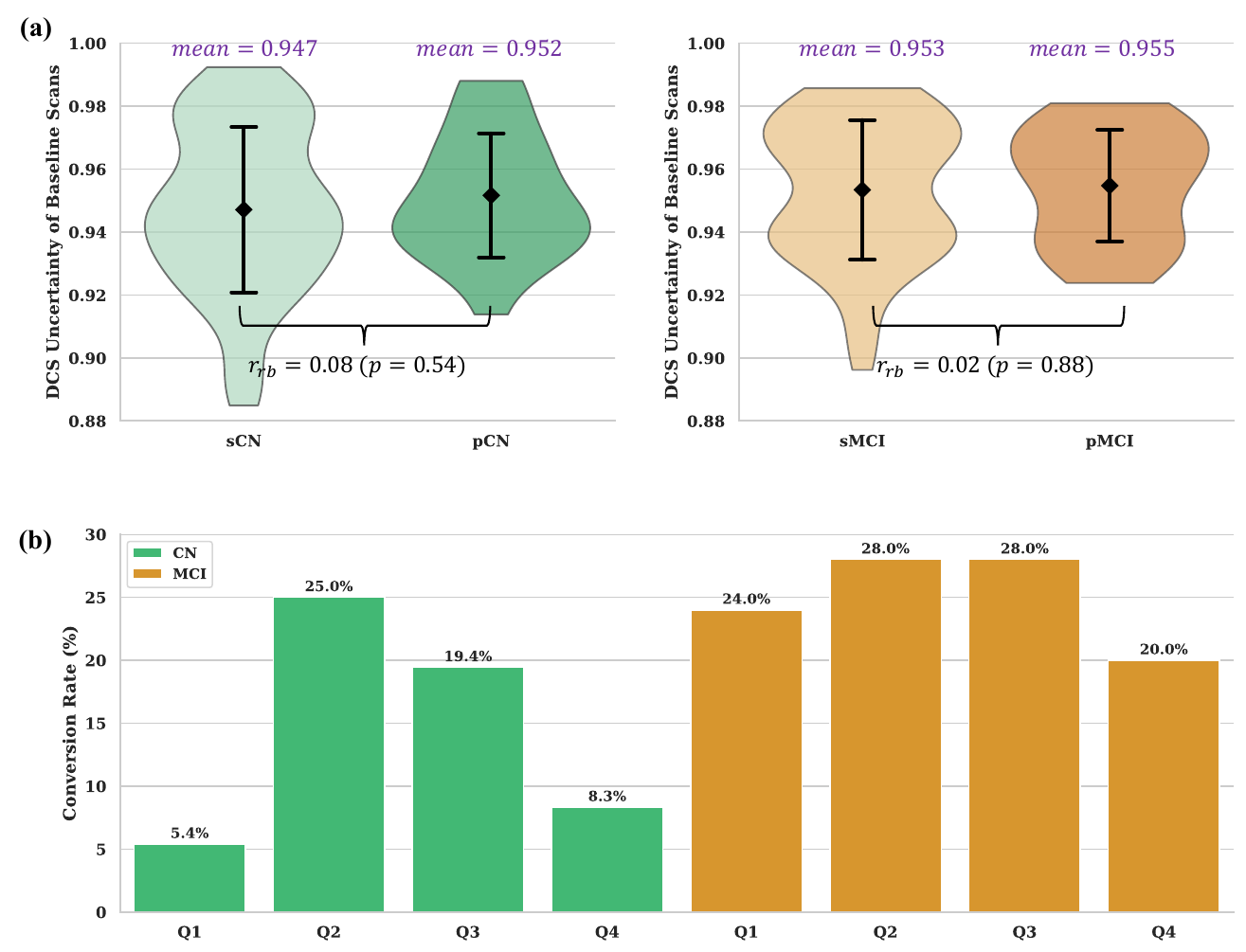}
  \caption{Analysis of posterior uncertainty associated with the Disease Continuum Score (DCS).
(a) Distributions of posterior uncertainty $\sigma$ derived from baseline scans for stable and progressive subjects in the CN and MCI cohorts.
Group differences were assessed using the Mann--Whitney $U$ test, with $r_{\mathrm{rb}}$ denoting the rank-biserial effect size.
(b) Observed conversion rates after stratifying subjects into quartiles according to baseline posterior uncertainty, from Q1 (lowest uncertainty) to Q4 (highest uncertainty), separately within the CN and MCI cohorts.
Unlike DCS, baseline posterior uncertainty shows neither a significant difference between stable and progressive subjects nor an ordered relationship with subsequent clinical conversion.
}
\label{fig:411_uncertainty}
\end{figure*}
The probabilistic formulation of DCP provides not only the estimated disease
position through DCS, but also the associated posterior uncertainty $\sigma$.
We therefore investigated whether posterior uncertainty derived from the baseline
scan, similar to DCS, contains information about subsequent clinical conversion.

Following the same grouping used in the conversion analysis, we compared baseline
posterior uncertainty between stable and progressive subjects.
The distributions were highly similar between the two groups in both cohorts.
For CN, the mean $\sigma$ was $0.947$ for sCN and $0.952$ for pCN
($U=1193.5$, $p=0.54$, $r_{rb}=0.08$), while for MCI, the corresponding values were
$0.953$ for sMCI and $0.955$ for pMCI
($U=957.5$, $p=0.88$, $r_{rb}=0.02$).
Consistently, stratification by quartiles of baseline posterior uncertainty did
not reveal an ordered relationship with subsequent conversion rate Figure \ref{fig:411_uncertainty}

In contrast to DCS, which showed clear associations with subsequent clinical
conversion, posterior uncertainty did not distinguish stable from progressive
subjects. These findings indicate that DCS and $\sigma$ characterize distinct
aspects of the probabilistic disease representation: DCS represents the estimated
position along the disease continuum, whereas $\sigma$ quantifies the uncertainty
associated with that position rather than serving as an indicator of future
clinical progression.

\section{Conclusion}
In this paper, we propose Disease Continuum Positioning (DCP), a Bayesian longitudinal framework for estimating the Disease Continuum Score (DCS) directly from longitudinal diffusion tensor imaging. By jointly integrating weak clinical supervision with longitudinal observations, DCP learns a probabilistic disease severity representation that continuously quantifies an individual's position along the Alzheimer's disease continuum together with its associated uncertainty. Extensive experiments demonstrate that the proposed framework consistently outperforms representative state-of-the-art methods in disease severity estimation, longitudinal disease characterization, and future disease conversion prediction. The resulting DCS provides a quantitative imaging-derived representation beyond conventional diagnostic labels and clinical scores, offering a promising tool for individualized disease monitoring and precision assessment of Alzheimer's disease. Although the proposed framework has demonstrated encouraging performance, its accuracy is inherently influenced by the availability of longitudinal follow-up data. Future studies with larger longitudinal cohorts and more complete follow-up visits are expected to further improve the estimation of disease severity and enable more accurate characterization of long-term disease trajectories.

\section*{Acknowledgment}
Part of this work used the Bridges-2 system, which is supported by NSF OAC-1928147 at the Pittsburgh Supercomputing Center (PSC).
We also acknowledge the UTRGV High Performance Computing Resource\footnote{\url{https://hpc.utrgv.edu/}}, supported by NSF grants 2018900 and IIS-2334389, and DoD grant W911NF2110169. 
Data used in preparation of this article were obtained from the Alzheimer’s Disease Neuroimaging Initiative (ADNI) database\footnote{\url{http://adni.loni.usc.edu}} funded by NIH grant U19AG024904. 
As such, the investigators within the ADNI contributed to the design and implementation of ADNI and/or provided data but did not participate in analysis or writing of this report. 
A complete listing of ADNI investigators can be found at \footnote{\url{http://adni.loni.usc.edu/wp-content/uploads/how_to_apply/ADNI_Acknowledgement_List.pdf}}.

\section*{Funding Declaration}
This study is partially supported by the National Institutes of Health (R33AG087888, R01AG092661, U01AG068057), AHA award 26AIREA1574568 and the National Science Foundation (CCF 2523787, IIS 2319450, IIS 2045848).

\bibliographystyle{unsrtnat}

\bibliography{cas-refs}






\end{document}